\documentclass[letterpaper]{article} % DO NOT CHANGE THIS
\usepackage{aaai25}  % DO NOT CHANGE THIS
\usepackage{times}  % DO NOT CHANGE THIS
\usepackage{helvet}  % DO NOT CHANGE THIS
\usepackage{courier}  % DO NOT CHANGE THIS
\usepackage[hyphens]{url}  % DO NOT CHANGE THIS
\usepackage{graphicx} % DO NOT CHANGE THIS
\usepackage{natbib}  % DO NOT CHANGE THIS AND DO NOT ADD ANY OPTIONS TO IT
\usepackage{caption} % DO NOT CHANGE THIS AND DO NOT ADD ANY OPTIONS TO IT
\usepackage{algorithm}
\usepackage{algorithmic}

\usepackage{booktabs}
\usepackage{color, colortbl}
\definecolor{mygray}{gray}{0.9}
\usepackage{amsfonts}
\usepackage{amsmath}
\usepackage{tikz} % 使用 tikz 包

\usepackage{newfloat}
\usepackage{listings}
\DeclareCaptionStyle{ruled}{labelfont=normalfont,labelsep=colon,strut=off} % DO NOT CHANGE THIS
\floatstyle{ruled}
\newfloat{listing}{tb}{lst}{}
\floatname{listing}{Listing}
\title{Pose Magic: Efficient and Temporally Consistent Human Pose Estimation\\ with a Hybrid Mamba-GCN Network}
\author{
    Xinyi Zhang\textsuperscript{\rm 1},
    Qiqi Bao\textsuperscript{\rm 2},
    Qinpeng Cui\textsuperscript{\rm 1},
    Wenming Yang\textsuperscript{\rm 1},
    Qingmin Liao\textsuperscript{\rm 1}\thanks{Corresponding author}
}
\affiliations{
    \textsuperscript{\rm 1}Tsinghua Shenzhen International Graduate School, Tsinghua University, Shenzhen, China\\
    \textsuperscript{\rm 2}Zhejiang University of Science \& Technology, Hangzhou, China\\
    xinyi-zh22@mails.tsinghua.edu.cn, nora919530829@163.com, 
    cqp22@mails.tsinghua.edu.cn,
    yang.wenming@sz.tsinghua.edu.cn,
    liaoqm@tsinghua.edu.cn
}

\usepackage{bibentry}
\begin{document}

\maketitle

\begin{abstract}
Current state-of-the-art (SOTA) methods in 3D Human Pose Estimation (HPE) are primarily based on Transformers. However, existing Transformer-based 3D HPE backbones often encounter a trade-off between accuracy and computational efficiency. To resolve the above dilemma, in this work, we leverage recent advances in state space models and utilize Mamba for high-quality and efficient long-range modeling. Nonetheless, Mamba still faces challenges in precisely exploiting local dependencies between joints. To address these issues, we propose a new attention-free hybrid spatiotemporal architecture named Hybr\textbf{i}d \textbf{Ma}mba-\textbf{GC}N (Pose Magic). This architecture introduces local enhancement with GCN by capturing relationships between neighboring joints, thus producing new representations to complement Mamba's outputs. By adaptively fusing representations from Mamba and GCN, Pose Magic demonstrates superior capability in learning the underlying 3D structure. To meet the requirements of real-time inference, we also provide a fully causal version. Extensive experiments show that Pose Magic achieves new SOTA results ($\downarrow 0.9 mm$) while saving $74.1\%$ FLOPs. In addition, Pose Magic exhibits optimal motion consistency and the ability to generalize to unseen sequence lengths.
\end{abstract}

\section{Introduction}

\begin{figure}
\centering
\includegraphics[width=1.0\linewidth]{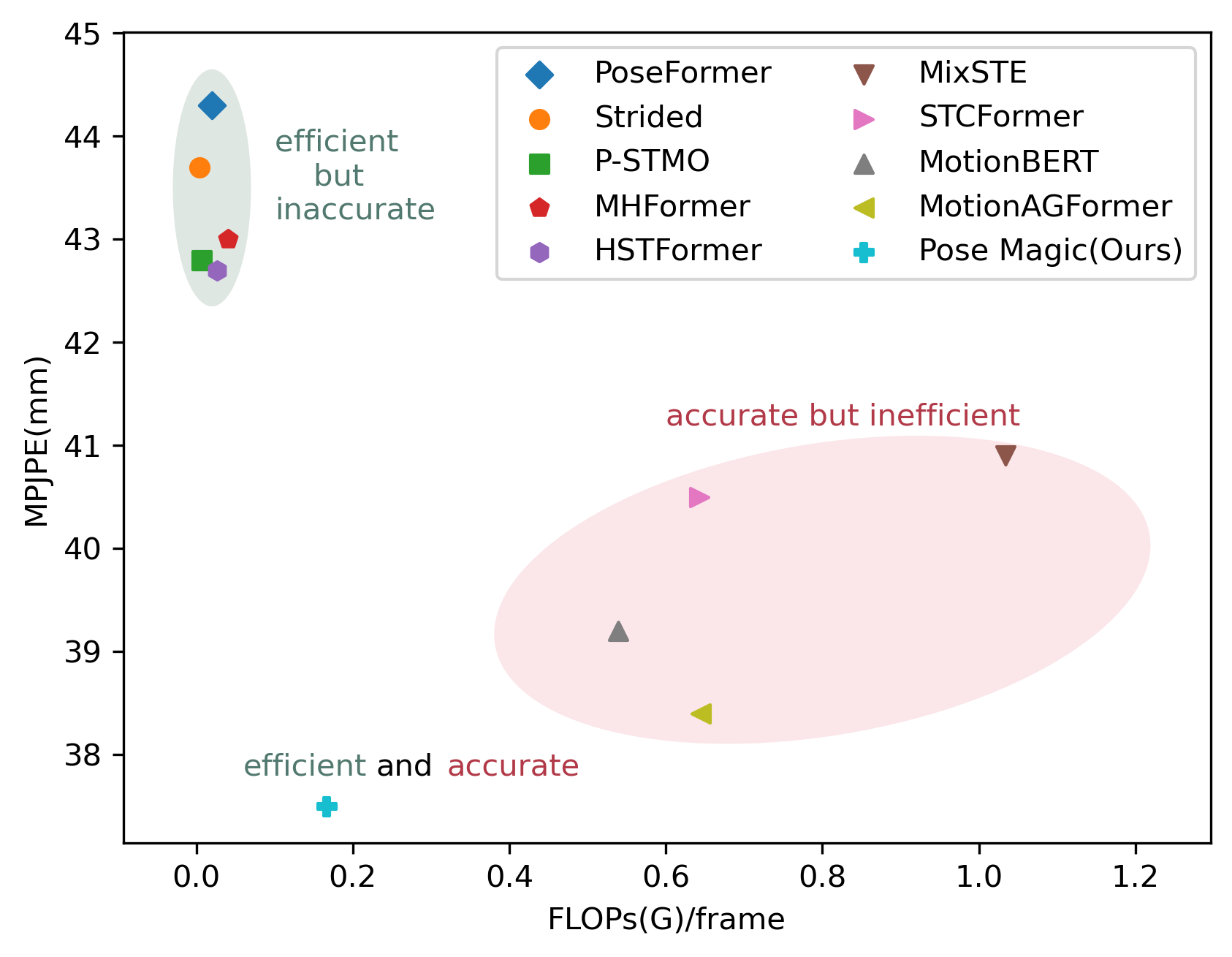}
\caption{Comparisons of transformer-based methods on Human3.6M ($\downarrow$). FLOPs/frame denotes floating point operations per output frame. The proposed Pose Magic attains superior results, while maintaining computational efficiency.}
\label{MPJPE_vs_FLOPs}
\end{figure}

Monocular 3D Human Pose Estimation (HPE) aims to capture positions of joints on the human skeleton in 3D space from images or videos. It is widely applied in action recognition \cite{peng2024navigating}, action correction and online coaching \cite{dittakavi2022pose}, and augmented/virtual reality \cite{yuan2023learning}. With these extensive applications, the demand for more accurate, computationally efficient and temporally consistent models continues to grow.

Impressive progress has been made in monocular 3D HPE \cite{zhu2023motionbert,mehraban2024motionagformer}. But it inherently remains an ill-posed problem due to depth ambiguity. To mitigate this issue in a single frame, some efforts have been made to comprehensively exploit the spatial and temporal relationships between joints contained in the input video. Current state-of-the-art (SOTA) methods \cite{li2022exploiting,li2022mhformer,zhang2022mixste,zhu2023motionbert} use Transformers \cite{vaswani2017attention} to capture spatiotemporal information from 2D pose sequences. Their outstanding performance is largely attributed to the powerful self-attention mechanism and the ability to capture long-range dependencies. However, despite the long-range receptive field, Transformer-based methods have high computational complexity (quadratic growth with the number of frames), making it challenging to deploy on devices with limited computational resources. Moreover, employing efficient techniques such as token pruning \cite{ma2022ppt,li2023hourglass} for 3D HPE often sacrifices a globally effective receptive field. These methods do not inherently resolve the trade-off between accuracy and computational efficiency.

Recently, achievements in State Space Models (SSMs), particularly Mamba \cite{gu2023mamba,liu2024vmamba,zhu2024vision}, have established them as an efficient and effective backbone for building deep networks. To address the above issue, \textbf{we propose a novel attention-free spatiotemporal architecture.} Specifically, both spatial and temporal Mamba blocks are included, which effectively learn rich spatiotemporal features. This new attention-free structure inherits the computational efficiency advantages of Mamba. Additionally, due to Mamba's inherent continuous-time formulation, the consistency and smoothness in pose estimation over time are improved.

However, standard Mamba still primarily accommodates long-range dependencies and pays less attention to local dependencies, which is detrimental to capturing human motion. Human movement inherently comprises local spatial and temporal dependencies \cite{mehraban2024motionagformer}. To address the above challenge, \textbf{we propose a hybrid spatiotemporal architecture} combining Mamba and Graph Convolutional Networks (GCNs) \cite{kipf2016semi}, named Hybr\textbf{i}d \textbf{Ma}mba-\textbf{GC}N (Pose Magic), to learn 3D pose representations from global to local. Specifically, Pose Magic consists of two streams: the Mamba stream and the GCN stream, which learn global and local information, respectively. In the GCN stream, GCN sequentially learns spatial and temporal relationships similar to the Mamba stream. We then employ adaptive fusion to aggregate features from both Mamba and GCN streams. By doing so, we ensure a balanced and comprehensive representation of human motion, thereby improving the accuracy of 3D HPE. Additionally, GCN has lower computational complexity, making our model still computationally efficient, as shown in Fig.\ref{MPJPE_vs_FLOPs}. We also introduce a fully causal version of Pose Magic, designed to predict each current timestep using only past and present frames, without forecasting future frames. This is crucial for real-time applications.

In summary, our main contributions are as follows:

\begin{itemize}
    \item We propose a novel attention-free hybrid network, Pose Magic. It leverages Mamba for high-quality and efficient long-range modeling. Additionally, GCN is utilized to enhance Mamba's performance by effectively capture local dependencies between joints. To the best of our knowledge, this is the first attempt to apply Mamba to 3D HPE tasks.
    \item We propose two versions of Pose Magic, achieved by designing the Bidirectional Magic Block and Unidirectional (causal) Magic Block, corresponding to offline and real-time inference, respectively.
    \item Extensive experiments on two popular benchmarks demonstrate that Pose Magic achieves state-of-the-art results while maintaining efficiency with fewer parameters and lower computational complexity.
    \item Pose Magic can model the natural smoothness of human motion, achieving optimal motion consistency and generalizing well to unseen sequence lengths.
\end{itemize}

\section{Related Work}
\subsection{3D Human Pose Estimation}
Based on the number of camera views used, 3D HPE can be categorized into monocular \cite{li2022mhformer,zhang2022mixste,zhu2023motionbert} and multi-view methods \cite{qiu2019cross,10.1145/3664647.3681265,zhang2024deep}. Due to the accessibility of a single camera in real-world scenarios, more attention has been paid to monocular 3D HPE, i.e., estimating 3D human pose from monocular images or videos. Mainstream monocular 3D HPE methods follow a two-stage paradigm. First, a plug-and-play 2D pose detector \cite{newell2016stacked,chen2018cascaded} is used to locate the 2D positions of joints. Then, the 2D pose is lifted to the 3D pose. In this work, we focus on the latter challenging step, known as the 2D-to-3D lifting process, following \citet{zhu2023motionbert,mondal2024hummuss}.

\subsection{Transformer-based Methods for 3D HPE}
Numerous studies have demonstrated Transformer \cite{vaswani2017attention} exhibits superior performance in 3D HPE. PoseFormer \cite{zheng20213d} is the first purely Transformer-based model, utilizing a spatial-temporal transformer to model joint relationships within each frame as well as temporal associations across frames. Subsequently, many studies have focused on enhancing spatiotemporal feature encoding. MixSTE \cite{zhang2022mixste} proposed learning distinct motion trajectories for each joint. HSTFormer \cite{qian2023hstformer} introduced a hierarchical Transformer encoder to capture multi-level spatiotemporal correlations from local to global. MotionBERT \cite{zhu2023motionbert} presented a dual-stream spatiotemporal Transformer that separately attends to long-range spatiotemporal relationships between stable and dynamic joints. To overcome self-occlusion and depth ambiguity, MHFormer \cite{li2022mhformer} learned spatiotemporal multi-hypothesis representations through Transformers.

However, performance gains come with significant computational overhead. Consequently, researchers have begun exploring efficient methods. Strided \cite{li2022exploiting} designed a cross-row Transformer encoder to aggregate redundant sequences, while \citet{shan2022p,zeng2022deciwatch,einfalt2023uplift} improved efficiency by uniform sampling of video sequences. In terms of token pruning, PPT \cite{ma2022ppt} selected important tokens based on attention scores. TCFormer \cite{zeng2022not} proposed a token clustering Transformer to merge tokens. In contrast to spatial reduction methods, Hourglass Tokenizer \cite{li2023hourglass} selected representative frame tokens in the temporal domain. However, these methods may sacrifice accuracy as they lose contextual cues.

\subsection{State Space Models}
Recent studies \cite{gu2021efficiently,gu2022parameterization,gupta2022diagonal,smith2022simplified} have explored SSMs as an effective alternative to Transformers for efficient modeling. For instance, \citet{gu2021efficiently,gu2022parameterization} proposed the S4 model and its diagonal version S4D, addressing issues of computational efficiency and long sequence dependencies. However, these models process all inputs in the same manner, which limits their data modeling capabilities. To enhance content-awareness, Mamba \cite{gu2023mamba} integrated time-varying parameters into the SSM framework and proposed a hardware-aware algorithm for efficient training and inference. Some researchers have applied SSMs to computer vision tasks. Vim \cite{zhu2024vision} introduced a bidirectional SSM block \cite{wang2022pretraining} for efficient and versatile visual representation, achieving performance comparable to ViT \cite{dosovitskiy2020image}. \citet{mondal2024hummuss} applied S4D to accelerate training and real-time inference in 3D HPE, albeit at the cost of sacrificing accuracy. In this work, we explore adapting Mamba to 3D HPE, effectively achieving a balance between accuracy and efficiency.

\subsection{Graph Convolutional Networks}
Human motion inherently involves local spatial and temporal dependencies \cite{mehraban2024motionagformer}. Therefore, it is crucial to model local dependencies. GCNs are computational efficient networks that specialize in local dependencies, achieving notable success in 3D HPE \cite{ci2019optimizing,zhao2019semantic,yu2023gla}. GLA-GCN \cite{yu2023gla} used global spatiotemporal representation and local joint representation to achieve a lower memory load while maintaining accuracy. MotionAGFormer \cite{mehraban2024motionagformer} integrated local spatial and temporal relationships using GCNs to complement the global information extracted by Transformers. However, existing methods do not consider temporal causality, which is impractical for real-time applications. In this work, we introduce local enhancement with GCN to complement the global outputs of Mamba. On this basis, we explore causal GCN methods by designing causal adjacency matrices.

\section{Method}
\subsection{Preliminaries}
\textbf{Selective Structured State Space Models.} SSMs \cite{gu2021efficiently,gu2023mamba} define a continuous system that maps a 1D sequence $x(t)\in \mathbb{R}$ to $y(t)\in \mathbb{R}$ through implicit latent states $h(t)\in \mathbb{R}^{N}$. This process is described by ordinary differential equations as follows:
\begin{equation}
\begin{aligned} 
h^{\prime}(t) & =\boldsymbol{A} h(t)+\boldsymbol{B} x(t),\\
y(t) & = \boldsymbol{C} h(t),
\end{aligned}
\label{ode}
\end{equation}
where $\boldsymbol{A}$, $\boldsymbol{B}$ and $\boldsymbol{C}$ are learned matrices. Instead of directly initializing $\boldsymbol{A}$ randomly, a popular strategy is to impose a diagonal structure on $\boldsymbol{A}$ \cite{gu2022parameterization,gupta2022diagonal,smith2022simplified}. 

To enhance computational efficiency, it is necessary to discretize continuous variables $\boldsymbol{A}$ and $\boldsymbol{B}$. The choice of discretization criteria is varied, with Zero-Order Hold \cite{iserles2009first} being a common approach.
\begin{equation}
    \overline{\boldsymbol{A}}=\exp (\Delta \boldsymbol{A}) \quad \overline{\boldsymbol{B}}=(\Delta \boldsymbol{A})^{-1}(\exp (\Delta \boldsymbol{A})-\boldsymbol{I}) \cdot \Delta \boldsymbol{B},
\end{equation}
where $\Delta$ represents the step size, and $\boldsymbol{I}$ is the identity matrix.

Thus, the discrete version of Eq.(\ref{ode}) can be written as:
\begin{equation}
    \begin{aligned} 
    h_{t} & =\overline{\boldsymbol{A}} h_{t-1}+\overline{\boldsymbol{B}} x_{t} \\ 
    y_{t} & =\boldsymbol{C} h_{t}.
    \end{aligned}
\end{equation}
This Linear Time-Invariant (LTI) discrete system allows efficient computation in recursive or convolution forms, scaling linearly or near-linearly with sequence lengths.

However, the selective state space model Mamba \cite{gu2023mamba} highlights that LTI limits the model's ability in data modeling. This is because matrices in SSMs remain unchanged regardless of input, thereby lacking content-based inference capabilities. Therefore, Mamba removes the constraint of LTI and introduces time-varying parameters, i.e.,
\begin{equation}
\begin{aligned}
& \quad \quad \quad \boldsymbol{B}=\operatorname{Linear}_{N}(x), \boldsymbol{C}=\operatorname{Linear}_{N}(x),\\
&\!\!\Delta\!=\!\operatorname{softplus}(\operatorname{Parameter}\!+\! \operatorname{Broadcast}_{D}(\operatorname{Linear}_{1}(x))), 
\end{aligned}
\end{equation}
where $\operatorname{Linear}_{d}$ is a parameterized projection layer projecting to dimension $d$, and $\operatorname{softplus}$ is an activation function.

However, the absence of LTI leads to a loss of equivalence with convolution, affecting training efficiency. To address this issue, Mamba introduces a hardware-aware algorithm, enabling parallelization. As a result, Mamba enables high-quality and efficient long-sequence dynamic modeling.

\textbf{Graph Convolutional Networks.} Unlike Mamba capturing global information, GCNs \cite{kipf2016semi} excel at aggregating local dependencies. A widely used GCN module \cite{luo2022learning} is defined as: 
\begin{equation}
\label{gcn}
\operatorname{GCN}(x)\!=\!\sigma\!\left(\!x\!+\!\operatorname{Norm}\!\left(\tilde{D}^{-\frac{1}{2}} \tilde{A} \tilde{D}^{-\frac{1}{2}} x W_{1}\!+\!x W_{2}\right)\right),
\end{equation}
where $\tilde{A}=A+\boldsymbol{I}$ represents the adjacency matrix with self-connections added. $\tilde{D} i i=\Sigma_j \tilde{A}_{ij}$ represents the summation of $\tilde{A}$ along its diagonal. $W_1$ and $W_2$ are trainable weight matrices. $\operatorname{Norm}(\cdot)$ and $\sigma(\cdot)$ denote Batch Normalization \cite{ioffe2015batch} and ReLU activation function \cite{glorot2011deep}, respectively. 

The key to GCNs lies in constructing the adjacency matrix $\tilde{A}$, which should be customized based on specific tasks.

\begin{figure}[t]
\centering
\includegraphics[width=0.8\linewidth]{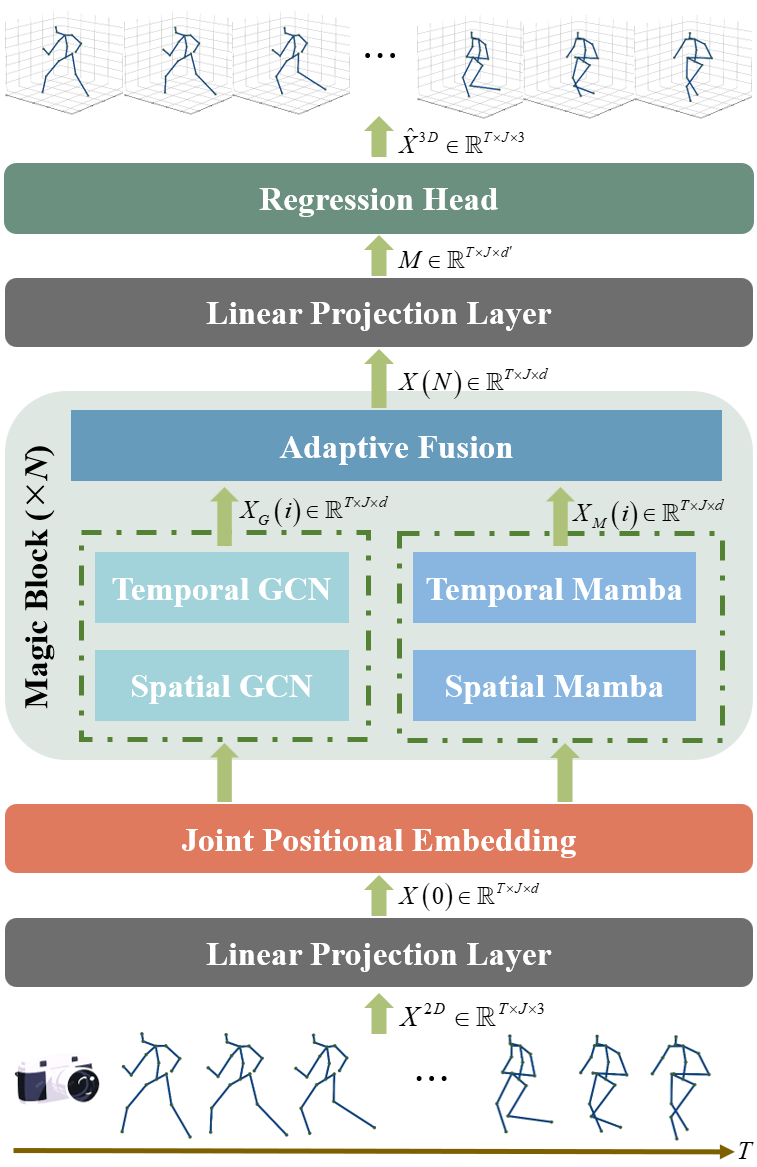}
\caption{Overview of Pose Magic. It consists of $N$ dual-stream Magic Blocks, with GCN capturing local information and Mamba capturing global information. Spatial GCN/Mamba models connections among joints within a frame, while the Temporal one tracks each joint's motion over time.}
\label{overal}
\end{figure}

\subsection{Overall Architecture}
To accurately lift 2D skeleton sequences to 3D pose sequences in a monocular setup, we propose the attention-free architecture Pose Magic. It employs Mamba and GCN to lift motion sequences, as shown in Fig.\ref{overal}.

Formally, given a monocular video of 2D joints with confidence scores $X^{2D}\in \mathbb{R}^{T\times J\times 3}$, our goal is to predict 3D positions $\hat{X}^{3D}\in \mathbb{R}^{T\times J\times 3}$. Here, $T$ and $J$ are the number of frames and joints, respectively. First, a linear projection layer is used to map each joint in every frame to a $d$-dimensional feature $X(0)\in \mathbb{R}^{T\times J\times d}$. Then, positional embedding $P_{pos}\in \mathbb{R}^{T\times J\times d}$ is added. Next, $N$ Magic Blocks effectively capture 3D structure of the skeleton sequence by computing $X(i)\in \mathbb{R}^{T\times J\times d} (i = 1,…,N)$. Subsequently, $X(N)$ is mapped to a higher dimension $M\in \mathbb{R}^{T\times J\times d'}$ using a linear layer and a $\operatorname{tanh}$ activation. Finally, a regression head is employed to output the 3D pose $\hat{X}^{3D}\in \mathbb{R}^{T\times J\times 3}$.

To ensure the temporal smoothness of the predicted 3D poses, we use a positional loss $\mathcal{L}_{3D}$ and a velocity loss $\mathcal{L}_v$ for supervision, i.e.,
\begin{equation}
    \mathcal{L}=\mathcal{L}_{3D}+\lambda \mathcal{L}_{v},
\end{equation}
where $\mathcal{L}_{3D}\!=\!\sum \limits_{t=1}^{T} \sum \limits_{j=1}^{J}\|\hat{X}^{3D}_{t, j}\!-\!X^{3D}_{t, j}\|$, $\mathcal{L}_v\!=\!\sum \limits_{t=2}^{T} \sum \limits_{j=1}^{J}\|\Delta \hat{X}^{3D}_{t, j}\!-\\
\Delta X^{3D}_{t, j}\|$. $\Delta \hat{X}^{3D}_{t}\!=\!\hat{X}^{3D}_{t}\!-\!\hat{X}^{3D}_{t-1}$ and $\Delta X^{3D}_{t}=X^{3D}_{t}-X^{3D}_{t-1}$. $\lambda$ is a balancing constant.

In the subsequent sections, we present the general architectures of bidirectional and unidirectional (causal) Magic Blocks, showing how to capture spatiotemporal information.

\subsection{Bidirectional Magic Block}
As shown in Fig.\ref{overal}, Magic Block comprises two streams: Mamba stream and GCN stream. Mamba stream leverages its high-quality and efficient long-range modeling capabilities to capture global dependencies. Meanwhile, GCN stream enhances the power of Mamba by effectively capturing local dependencies between joints. This dual-stream architecture ensures comprehensive and accurate 3D pose estimation. Specifically, Spatial Mamba/GCN treats different joints as individual tokens, effectively capturing the structural relationships of joints within a frame. Temporal Mamba/GCN considers each frame as a single token, thereby capturing the motion trajectory of joints over time. Finally, features captured by both streams are adaptively fused.

\begin{figure}[t]
    \begin{minipage}[t]{0.5\linewidth}
        \centering
        \includegraphics[width=1.0\textwidth]{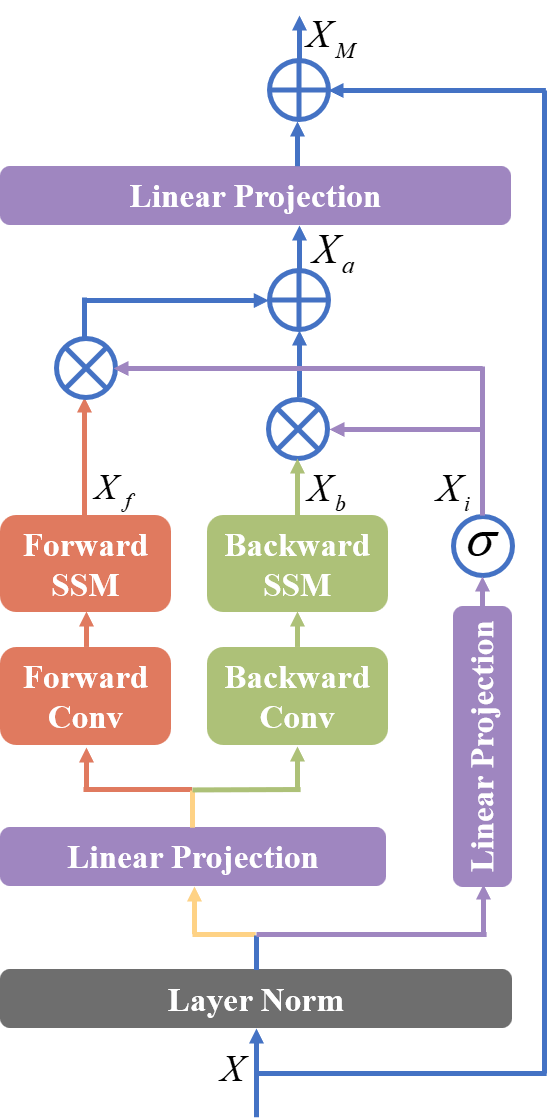}
        \centerline{\textbf{(a) Bidirectional Mamba}}
    \end{minipage}%
    \begin{minipage}[t]{0.5\linewidth}
        \flushright
        \includegraphics[width=0.672\textwidth]{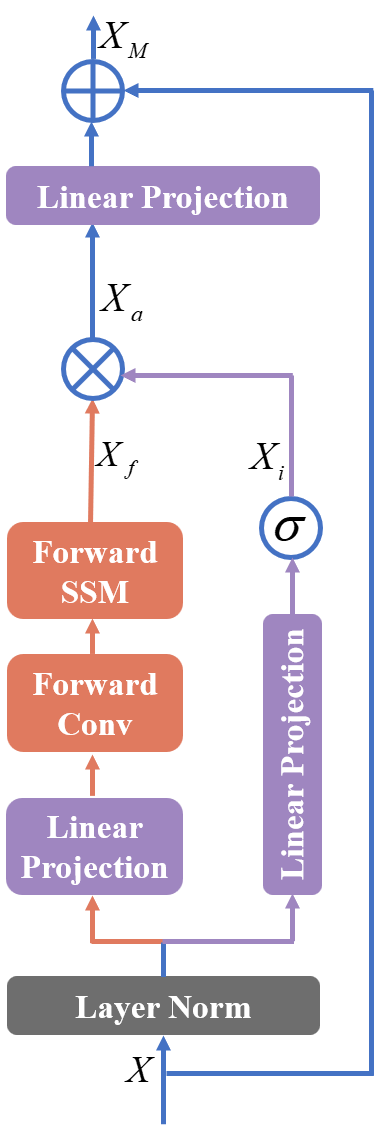}
        \centerline{\textbf{(b) Unidirectional Mamba}}
    \end{minipage}
    \caption{Different Mamba structures. (a) Bidirectional: process information forward, backward and independently. (b) Unidirectional: process information forward and independently. Here, current information only relates to present and past data, making it suitable for real-time applications.}
\label{mamba}
\end{figure}

\textbf{Mamba Stream.} To bidirectionally capture spatiotemporal information, we adopt the structure of \citet{zhu2024vision}, as shown in Fig.\ref{mamba}(a). This stream has three parallel processing paths: forward, backward, and independent.

Formally, given an input $X\in \mathbb{R}^{B\times L \times d}$, where $B$ is the batch size, $L$ is the length of the sequence and $d$ is the dimension. First, information is processed forward and backward along the sequence dimension, respectively:
\begin{equation}
    X_f = \operatorname{SSM}_f(\sigma(\operatorname{Norm}_l(X)W_{p1}W_f)),
\end{equation}
\begin{equation}
    X_b = \operatorname{flip}(\operatorname{SSM}_b(\sigma(\operatorname{flip}(\operatorname{Norm}_l(X)W_{p1})W_b))).
\end{equation}

The final path processes information independently:
\begin{equation}
X_i = \sigma(\operatorname{Norm}_l(X)W_{p2}),
\end{equation}
where $\operatorname{Norm}_l(\cdot)$ represents Layer Normalization, $\sigma(\cdot)$ is the GELU activation function \cite{hendrycks2016gaussian}, and $\operatorname{flip}(\cdot)$ indicates flipping along the sequence dimension. $W_{p1},W_{p2},W_f,W_b\in \mathbb{R}^{d \times d}$ are learnable matrices.

Next, information from the forward, backward and independent paths is aggregated via multiplicative gating:
\begin{equation}
\label{aggregation}
X_a = X_f\odot X_i+ X_b\odot X_i,
\end{equation}
where $\odot$ denotes a Hadamard Product.

Finally, a skip connection is added to compute the output:
\begin{equation}
X_M = X+X_a W_{p3},
\end{equation}
where $W_{p3}\in \mathbb{R}^{d \times d}$ is a learnable matrix.

\begin{figure}[H]
    \begin{minipage}[t]{0.5\linewidth}
        \centering
        \includegraphics[width=0.837\textwidth]{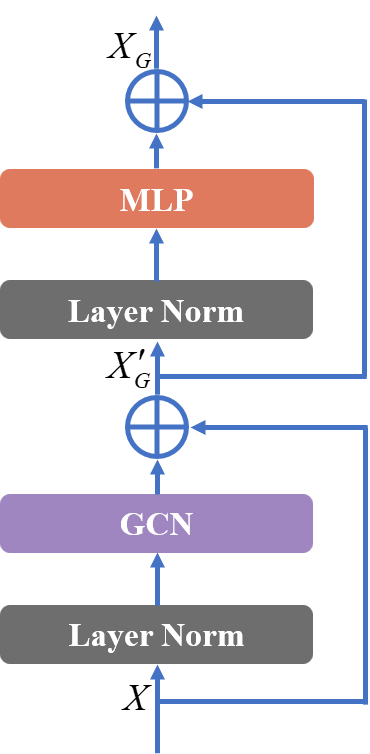}
        \centerline{\textbf{(a) GCN}}
    \end{minipage}%
    \begin{minipage}[t]{0.5\linewidth}
        \centering
        \includegraphics[width=0.95\textwidth]{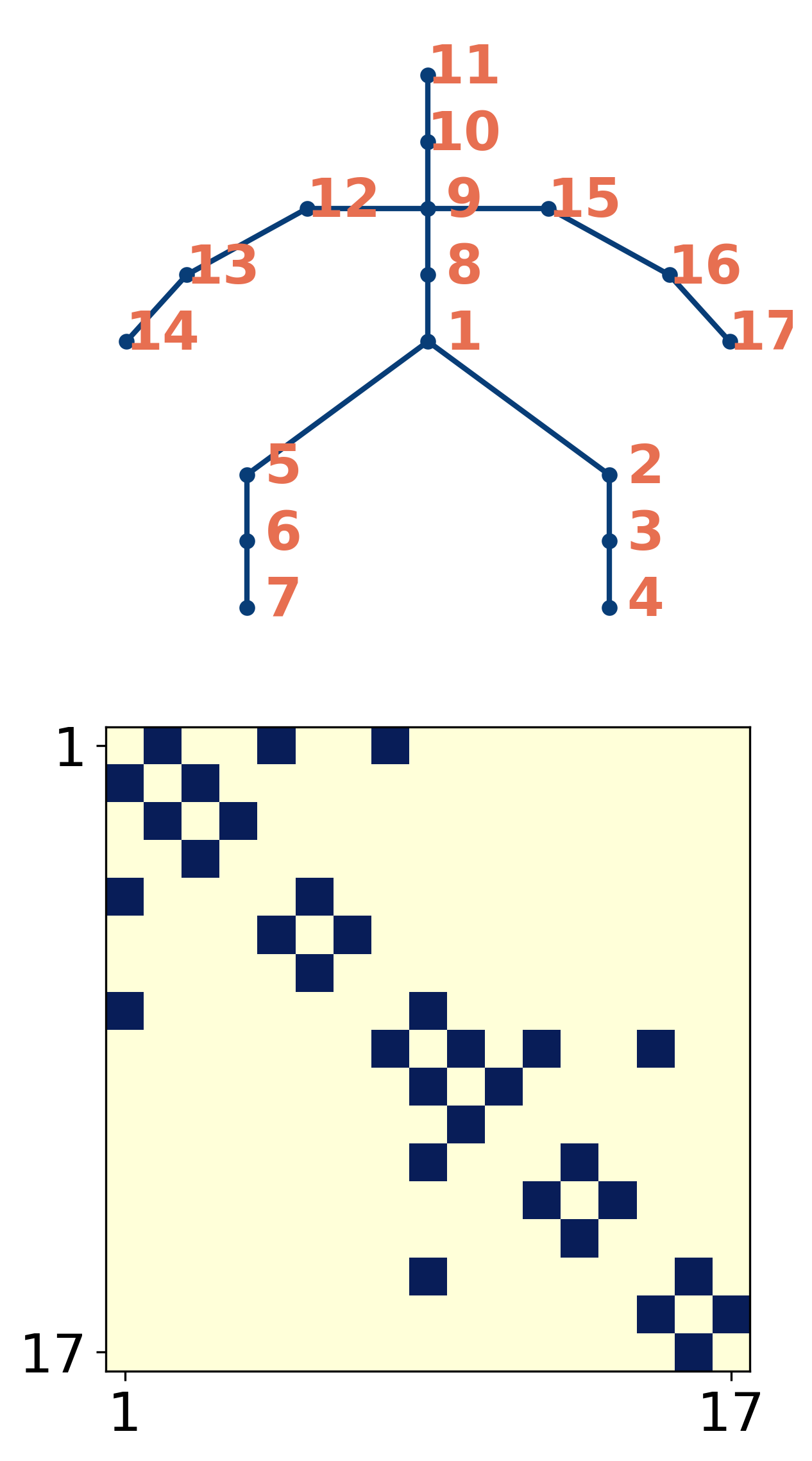}
        \centerline{\textbf{(b) Spatial GCN}}
    \end{minipage}\\
    \begin{minipage}[t]{1.0\linewidth}
        \centering
        \includegraphics[width=1\textwidth]{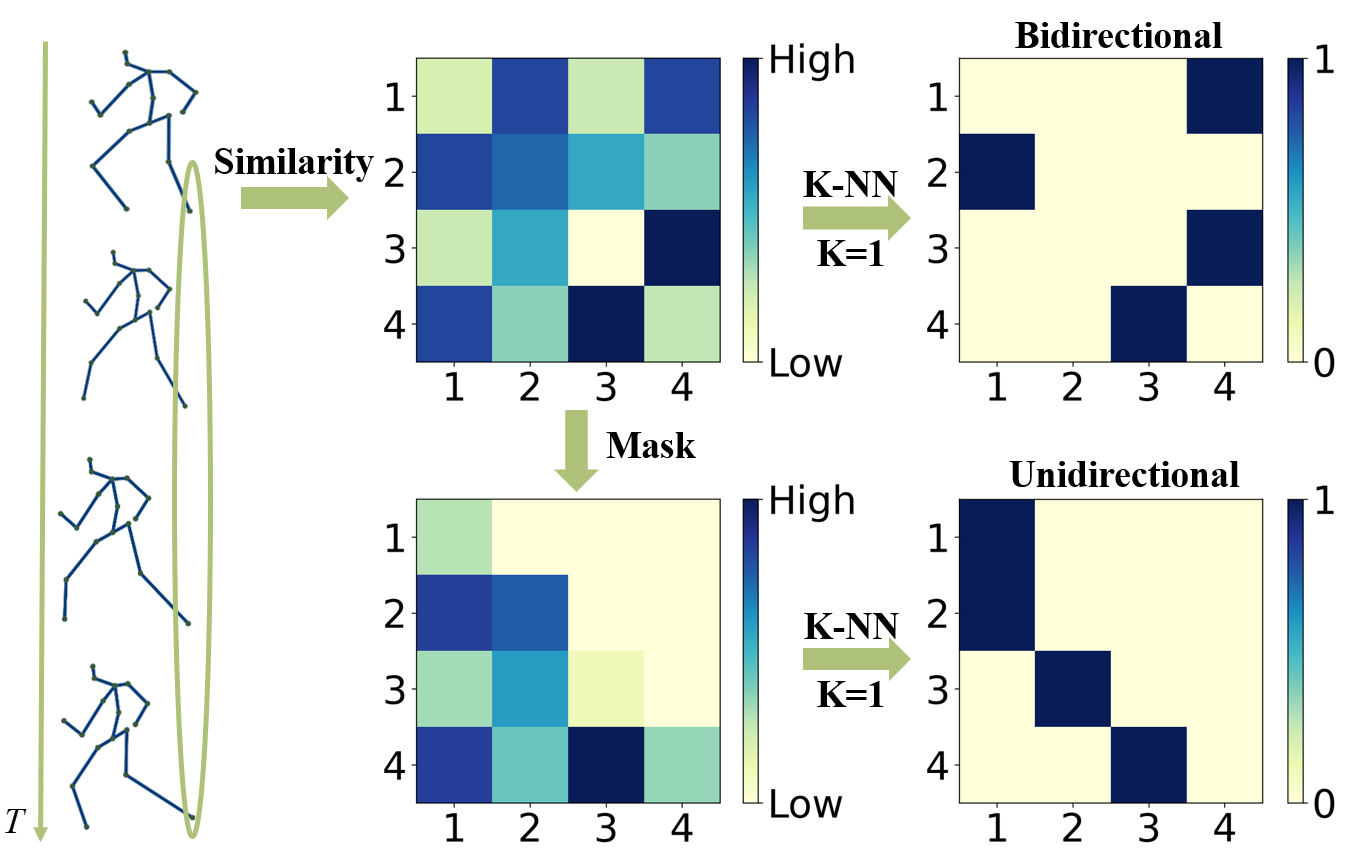}
        \centerline{\quad \textbf{(c) Adjacency matrix in Temporal GCN}}
    \end{minipage}
    \caption{(a) GCN structure. (b) Spatial GCN uses the Human3.6M skeleton as the adjacency matrix. (c) Temporal GCN uses K-NN for connection edges based on joint similarity across frames. After K-NN, each row connects to $K$ columns. Top: bidirectional adjacency matrix. Bottom: unidirectional adjacency matrix: K-NN after a causal mask.}
\label{GCN}
\end{figure}

\textbf{GCN Stream.} Unlike Mamba stream aggregating global information, GCN stream focuses on local spatial and temporal relationships to incorporate pose-specific priors. Our GCN structure is shown in Fig.\ref{GCN}(a). Given $X\in \mathbb{R}^{B \times L \times d}$, the output $X_G$ is obtained by
\begin{equation}
\begin{aligned}
X_G' &= X+\operatorname{GCN}(\operatorname{Norm}(X)), \\
X_G &= X_G' + \operatorname{MLP}(\operatorname{Norm}(X_G')),
\end{aligned}
\end{equation}
where $\operatorname{GCN}(\cdot)$ is defined in Eq.(\ref{gcn}).

As mentioned in \textit{Preliminaries}, adjacency matrices in GCNs should be customized based on specific requirements. In Spatial GCN, joint topology is chosen as the adjacency matrix (Fig.\ref{GCN}(b)). For Temporal GCN, we first calculate the similarity of individual joints across different frames:
\begin{equation}
\operatorname{Sim}(X_{t_i},X_{t_j})=(X_{t_i})^T X_{t_j}.
\end{equation}
Then, $k$ nearest neighbors are selected as connected nodes in the temporal graph (Fig.\ref{GCN}(c)). In a bidirectional setup, the model can access information from the entire sequence, thus a node at a certain frame may relate to past frames as well as future frames.

\textbf{Adaptive Fusion.} Following \citet{mehraban2024motionagformer}, we employ adaptive fusion to aggregate features extracted from Mamba and GCN streams:
\begin{equation}
\label{fusion}
\begin{aligned}
&\! X(i) = \alpha_M(i) \odot X_M (i-1) + \alpha_G (i) \odot X_G(i-1),\\
& \alpha_M(i), \alpha_G(i) \!=\!\operatorname{softmax}(W [X_M(i\!\!-\!\!1),X_G(i\!\!-\!\!1)]),
\end{aligned}
\end{equation}
where $X(i)$ denotes feature embeddings extracted at depth $i$, $X_M(i-1)$ and $X_G(i-1)$ represent features extracted from Mamba and GCN streams respectively at depth $i-1$. $W$ is a learnable matrix.

\subsection{Unidirectional Magic Block}
To ensure the temporal causality, we introduce the Unidirectional Magic Block, which also includes the branching and fusion of Mamba and GCN.

\textbf{Mamba Stream.} As shown in Fig.\ref{mamba}(b), for an input $X\in \mathbb{R}^{B\times L \times d}$, Unidirectional Mamba combines information along the sequence dimension $L$ but only learns in the forward and independent directions. In this case, Eq.(\ref{aggregation}) is rewritten to aggregate the forward and independent paths:
\begin{equation}
X_a = X_f\odot X_i.
\end{equation}

\textbf{GCN Stream.} The structure of Unidirectional GCN remains as shown in Fig.\ref{GCN}(a), but its temporal adjacency matrix undergoes some changes. Due to the causal nature of the model, which can only observe present and past data, the Unidirectional adjacency matrix is obtained by applying a causal mask to the similarity matrix, followed by filtering via K-nearest neighbors (K-NN). Specifically, we zero out similarities corresponding to frames occurring after the current time. This prevents K-NN from selecting temporal connections from future frames.

\textbf{Adaptive Fusion.} Features extracted from Mamba and GCN streams are also aggregated according to Eq.(\ref{fusion}).

\section{Experiments}
\subsection{Datasets and Evaluation Metrics}
Experiments are conducted on widely used 3D HPE benchmark datasets, including Human 3.6M \cite{ionescu2013human3} and MPI-INF-3DHP \cite{mehta2017monocular}.

\textbf{Human3.6M (H3.6M)} is one of the largest indoor motion capture datasets, consisting of 3.6 million frames from 11 actors across 15 scenarios. Following \citet{cai2024disentangled}, our model is trained on subjects S1, S5, S6, S7 and S8, and tested on subjects S9 and S11. We evaluate our single-view temporal 3D HPE model using three metrics: Mean Per Joint Position Error (MPJPE, $mm$) \cite{pavllo20193d} for joint accuracy, Mean Per Joint Velocity Error (MPJVE, $mm/s$) and Acceleration Error (ACC-ERR, $mm/s^2$) \cite{mehraban2024motionagformer} for temporal consistency and smoothness, which are essential for video applications. MPJVE and ACC-ERR correspond to the MPJPE of the first and second derivatives of 3D pose sequences.

\textbf{MPI-INF-3DHP (3DHP)} is a large dataset collected in various indoor and outdoor environments. It includes over 1.3 million frames, capturing 8 actors performing 8 different activities. Following \citet{shan2023diffusion}, MPJPE, Percentage of Correct Keypoint (PCK) within 150 mm, and the Area Under the Curve (AUC) are reported as evaluation metrics.

\subsection{Implementation Details}
\textbf{Model Variants.} Based on whether causality is considered, we build two different models: a bidirectional model and a causal model. The difference lies in the Temporal Magic Block: bidirectional model uses a Bidirectional Magic Block, while causal model uses a Unidirectional Magic Block. Both models use a bidirectional Spatial Magic Block. For all experiments, the number of layers is $N = 26$, with a hidden dimension of $d = 128$, a motion semantic dimension of $d'=512$, and the number of temporal neighbors in the GCN stream is $k = 2$.

\textbf{Experimental Settings.} Our model is implemented using PyTorch \cite{paszke2017automatic} and trained on two GeForce RTX 3090 GPUs. Following \citet{zhu2023motionbert}, horizontal flip augmentation is applied during both training and testing. For training, each mini-batch consists of 8 sequences. The AdamW optimizer \cite{loshchilov2017decoupled} is utilized for 90 epochs with a weight decay of $0.01$. The initial learning rate is $8e^{-4}$, with an exponential decay schedule and a decay factor of $0.99$. To ensure a fair comparison with previous studies \cite{zhu2023motionbert,mehraban2024motionagformer}, detected 2D poses from an off-the-shelf 2D pose detector \cite{newell2016stacked} are used for H3.6M, while ground truth 2D poses are used for 3DHP.

\subsection{Comparison with the State-of-the-art Methods}

\begin{table}[t]
\begin{center}
\scalebox{0.72}{
\begin{tabular}{l|cccc}
\hline
Method &$T$ &Params &FLOPs &MPJPE $\downarrow$
\\ \hline \hline
\multicolumn{5}{c}{Causal} \\
\hline
MotionBERT-scatch \cite{zhu2023motionbert}$^*$ &243	&16.00 &131.09	&44.7\\
\cite{mehraban2024motionagformer}$^*$ &243	&19.00	&156.63	&\textcolor{blue}{42.6}\\
\hline
\rowcolor{mygray}
Ours &243  &14.21	&40.58	&\textcolor{red}{41.7}  \\ 
\hline \hline
\multicolumn{5}{c}{Bidirectional} \\
\hline
PoseFormer \cite{zheng20213d} &81 &9.60 &1.63 &44.3 \\
\cite{einfalt2023uplift} &351 &10.36	&1.00  &44.2 \\
Strided \cite{li2022exploiting} &351 &4.35 &1.60 &43.7 \\
MHFormer \cite{li2022mhformer} &351 &31.52 &14.15 &43.0 \\
TPC w. MHFormer \cite{li2023hourglass} &351	&31.52	&8.22	&43.0\\
P-STMO \cite{shan2022p} &243 &7.01 &1.74 &42.8 \\
HSTFormer \cite{qian2023hstformer} &81 &22.72	&2.12	&42.7\\
HDFormer \cite{chen2023hdformer} &96	&3.70	&-	&42.6\\
HoT w. MixSTE \cite{li2023hourglass} &243	&35.00	&167.52	 &41.0\\
MixSTE \cite{zhang2022mixste} &243	&33.78	&277.25	&40.9\\
STCFormer \cite{tang20233d} &243	&18.93	&156.22	 &40.5\\
TPC w. MixSTE \cite{li2023hourglass} &243	&33.78	&251.29	 &39.9\\
MotionBERT-scatch \cite{zhu2023motionbert} &243	&16.00	&131.09	 &39.2\\
HoT w. MotionBERT \cite{li2023hourglass} &243	&16.35	&63.21	&39.8\\
TPC w. MotionBERT \cite{li2023hourglass} &243 &16.00	 &91.38	&39.0\\
\cite{mehraban2024motionagformer} &243	&19.00	&156.63	&\textcolor{blue}{38.4}\\
\hline
\rowcolor{mygray}
Ours &243 &14.42 &40.58 &\textcolor{red}{37.5}  \\ 
\hline
\end{tabular}
}
\end{center}
\caption{Comparison of parameters (M), FLOPs (G) and MPJPE with Transformer-based methods on Human3.6M. $T/^{*}$ denote the number of frames / our re-implementation. \textcolor{red}{Red}: Best. \textcolor{blue}{Blue}: Second best.} 
\label{table:h36m}
\end{table}

\begin{table}[t]
\begin{center}
\scalebox{0.72}{
\begin{tabular}{l|cccc}
\hline
Method & $T$ & MPJPE$\downarrow$ & PCK$\uparrow$ & AUC$\uparrow$ \\ \hline \hline
\multicolumn{5}{c}{Causal} \\
\hline
MotionBERT-scatch \cite{zhu2023motionbert} & 243 & 25.4 & 97.9 & 85.2 \\
\cite{mondal2024hummuss}(scatch) & 243 & 24.6 & 98.2 & \textcolor{blue}{85.6}\\
\cite{mehraban2024motionagformer}$^*$ & 81 &\textcolor{blue}{18.7}  &\textcolor{blue}{98.4} &85.0  \\
\hline
\rowcolor{mygray}
Ours &81  &\textcolor{red}{16.1} &\textcolor{red}{99.1}  &\textcolor{red}{86.7}  \\ 
\hline \hline
\multicolumn{5}{c}{Bidirectional} \\
\hline
PoseFormer \cite{zheng20213d} & 81  & 77.1 & 88.6 & 56.4 \\ 
TPC w. MHFormer \cite{li2023hourglass} &9	&58.4	&94.0	&63.3\\
MHFormer \cite{li2022mhformer} & 9  & 58.0 & 93.8 & 63.3 \\ 
MixSTE \cite{zhang2022mixste} & 27  & 54.9 & 94.4 & 66.5\\ 
HoT w. MixSTE \cite{li2023hourglass} &27	&53.2	&94.8	&66.5\\
\cite{einfalt2023uplift} & 81 &  46.9 & 95.4 & 67.6\\
HSTFormer \cite{qian2023hstformer} & 81  & 41.4& 97.3 &71.5\\
HDFormer \cite{chen2023hdformer} & 96  & 37.2 & 98.7 & 72.9\\
P-STMO \cite{shan2022p} & 81 & 32.2 & 97.9 & 75.8\\
PoseFormerV2 \cite{zhao2023poseformerv2} & 81  & 27.8 & 97.9 & 78.8\\
GLA-GCN \cite{yu2023gla} & 81 & 27.7 & 98.5 & 79.1\\
STCFormer \cite{tang20233d} & 81 & 23.1 & 98.7 & 83.9\\
\cite{mondal2024hummuss}(scatch) & 243 & 18.7 & \textcolor{blue}{99.0} & 87.1 \\
MotionBERT-scatch \cite{zhu2023motionbert} & 243  & 18.2 & \textcolor{red}{99.1} & \textcolor{red}{88.0} \\
\cite{mehraban2024motionagformer} & 81 & \textcolor{blue}{16.2} & 98.2 & 85.3 \\
\hline
\rowcolor{mygray}
%Ours &81 &\textcolor{red}{13.4}  &\textcolor{red}{99.4} &\textcolor{red}{88.4}   
Ours &81 &\textcolor{red}{14.7} &98.8 &\textcolor{blue}{87.6}\\ 
\hline
\end{tabular}
}
\end{center}
\caption{Comparison on MPI-INF-3DHP. $T$ denotes the number of input frames. $^{*}$ indicates our re-implementation. \textcolor{red}{Red}: Best. \textcolor{blue}{Blue}: Second best.}
\label{tab:3dhp}
\end{table}

\textbf{Results on Human3.6M.} Current SOTA performance on H3.6M is achieved by Transformer-based architectures, but these come with high computational complexity. We categorize methods based on causality and compare our approach with them in Table \ref{table:h36m}. In bidirectional settings, Pose Magic not only outperforms previous Transformer-based methods but also significantly reduces computational costs. Specifically, our Pose Magic achieves $37.5mm$ in MPJPE, improving accuracy by $0.9mm$ compared to the previous SOTA \cite{mehraban2024motionagformer}, while saving $74.1\%$ in FLOPs and improving Params by $4.58G$. Causal methods are more suitable for real-time scenarios where future frame information is unavailable. For strong baselines, we also train their non-causal variants. It can be observed that Pose Magic also outperforms existing methods in causal settings.

\textbf{Results on MPI-INF-3DHP.} We further evaluate our method on 3DHP, as shown in Table \ref{tab:3dhp}. Under causal and non-causal settings, our method with $T= 81$ consistently outperforms other approaches across all metrics, demonstrating its effectiveness in indoor and outdoor scenarios.

\subsection{Temporal Consistency and Smoothness}

\begin{table*}[t]
\begin{center}
\scalebox{0.7}{
\begin{tabular}{l|ccccccccccccccc|c}
\hline
Method   
&Dir. & Disc.& Eat & Greet & Phone & Photo & Pose & Purch. & Sit & SitD. & Smoke & Wait &  WalkD. & Walk  & WalkT. & Avg. \\
\hline \hline
\multicolumn{16}{c}{Causal}  \\
\hline
MotionBERT-scatch \cite{zhu2023motionbert}$^*$ &2.6 &2.9 &2.1 &2.9 &2.1 &2.8 &2.6 &2.9 &1.7 &2.5 &2.1 &2.0 &3.6 &2.6 &2.3 &2.5 \\
\cite{mehraban2024motionagformer}$^*$ &\textcolor{blue}{2.3} &\textcolor{blue}{2.5} &\textcolor{red}{1.8} &\textcolor{blue}{2.6} &\textcolor{blue}{1.8} &\textcolor{blue}{2.4} &\textcolor{blue}{2.2} &\textcolor{blue}{2.6} &\textcolor{red}{1.4} &\textcolor{blue}{2.1} &\textcolor{blue}{1.8} &\textcolor{blue}{1.7} &\textcolor{blue}{3.2} &\textcolor{blue}{2.4} &\textcolor{blue}{2.1} &\textcolor{blue}{2.2} \\
\hline
\rowcolor{mygray}
Ours &\textcolor{red}{2.2} &\textcolor{red}{2.4} &\textcolor{red}{1.8} &\textcolor{red}{2.5} &\textcolor{red}{1.7} &\textcolor{red}{2.2} &\textcolor{red}{2.1} &\textcolor{red}{2.5} &\textcolor{red}{1.4} &\textcolor{red}{2.0} &\textcolor{red}{1.7}  &\textcolor{red}{1.6} &\textcolor{red}{3.0} &\textcolor{red}{2.3} &\textcolor{red}{2.0} &\textcolor{red}{2.1} \\ 
\hline \hline
\multicolumn{16}{c}{Bidirectional}  &\\
\hline
PoseFormer \cite{zheng20213d} &3.2 &3.4 &2.6 &3.6 &2.6 &3.0 &2.9 &3.2 &2.6 &3.3 &2.7 &2.7 &3.8 &3.2 &2.9  & 3.1 \\ 
VPose \cite{pavllo20193d} & 3.0 &3.1 &2.2 &3.4 &2.3 &2.7 &2.7 &3.1 &2.1 &2.9 &2.3 &2.4 &3.7 &3.1 &2.8  &2.8\\
Trajectory Pose \cite{lin2019trajectory} &2.7 &2.8 &2.1 &3.1 &2.0 &2.5 &2.5 &2.9 &1.8 &2.6 &2.1 &2.3 &3.7 &2.7 &3.1  &2.7\\
Anatomy3D \cite{chen2021anatomy} &2.7 &2.8 &2.0 &3.1 &2.0 &2.4 &2.4 &2.8 &1.8 &2.4 &2.0 &2.1 &3.4 &2.7 &2.4 &2.5 \\
MHFormer \cite{li2022mhformer} &2.6 &2.7 &1.9 &2.8 &1.9 &2.3 &2.3 &2.6 &1.7 &2.4 &2.0 &2.1 &3.2 &2.7 &2.3  &2.4\\
MHFormer++ \cite{li2023multi} &2.5 &2.6 &1.9 &2.8 &1.9 &2.2 &2.3 &2.6 &1.7 &2.4 &1.9 &2.0 &3.1 &2.5 &2.2  & 2.3\\
MixSTE \cite{zhang2022mixste} & 2.5 &2.7 &1.9 &2.8 &1.9 &2.2 &2.3 &2.6 &1.6 &2.2 &1.9 &2.0 &3.1 &2.6 &2.2 &2.3\\
MotionBERT-scatch \cite{zhu2023motionbert} &1.8 &2.1 &\textcolor{blue}{1.5} &\textcolor{blue}{2.0} &\textcolor{blue}{1.5} &\textcolor{blue}{1.9} &\textcolor{blue}{1.8} &2.1 & 1.2 &1.8 &1.5 &\textcolor{blue}{1.4} &2.6 &\textcolor{blue}{2.0} &\textcolor{blue}{1.7} &\textcolor{blue}{1.8} \\
\cite{mehraban2024motionagformer} &\textcolor{blue}{1.8} &\textcolor{blue}{2.0} & \textcolor{red}{1.4} &\textcolor{blue}{2.0} &\textcolor{blue}{1.5} &2.0 &\textcolor{blue}{1.8} &\textcolor{blue}{2.0} &\textcolor{red}{1.1} &\textcolor{blue}{1.7} &\textcolor{blue}{1.4} &\textcolor{blue}{1.4} &\textcolor{blue}{2.5} & \textcolor{blue}{2.0} &\textcolor{blue}{1.7} &\textcolor{blue}{1.8}\\
\hline
\rowcolor{mygray}
Ours &\textcolor{red}{1.6} &\textcolor{red}{1.8} &\textcolor{red}{1.4} &\textcolor{red}{1.8} &\textcolor{red}{1.4} &\textcolor{red}{1.6} &\textcolor{red}{1.6} &\textcolor{red}{1.9} &\textcolor{red}{1.1} &\textcolor{red}{1.6} &\textcolor{red}{1.3} &\textcolor{red}{1.3}  &\textcolor{red}{2.3} & \textcolor{red}{1.9} &\textcolor{red}{ 1.6}  &\textcolor{red}{1.6}  \\ 
\hline
\end{tabular}
}
\end{center}
\caption{Comparison results of MPJVE on Human3.6M. $^{*}$ indicates our re-implementation. \textcolor{red}{Red}: Best. \textcolor{blue}{Blue}: Second best.}
\label{mpjve}
\end{table*}

\begin{figure}[t]
    \begin{minipage}[t]{0.52\linewidth}
        \centering
        \includegraphics[width=1.0\textwidth]{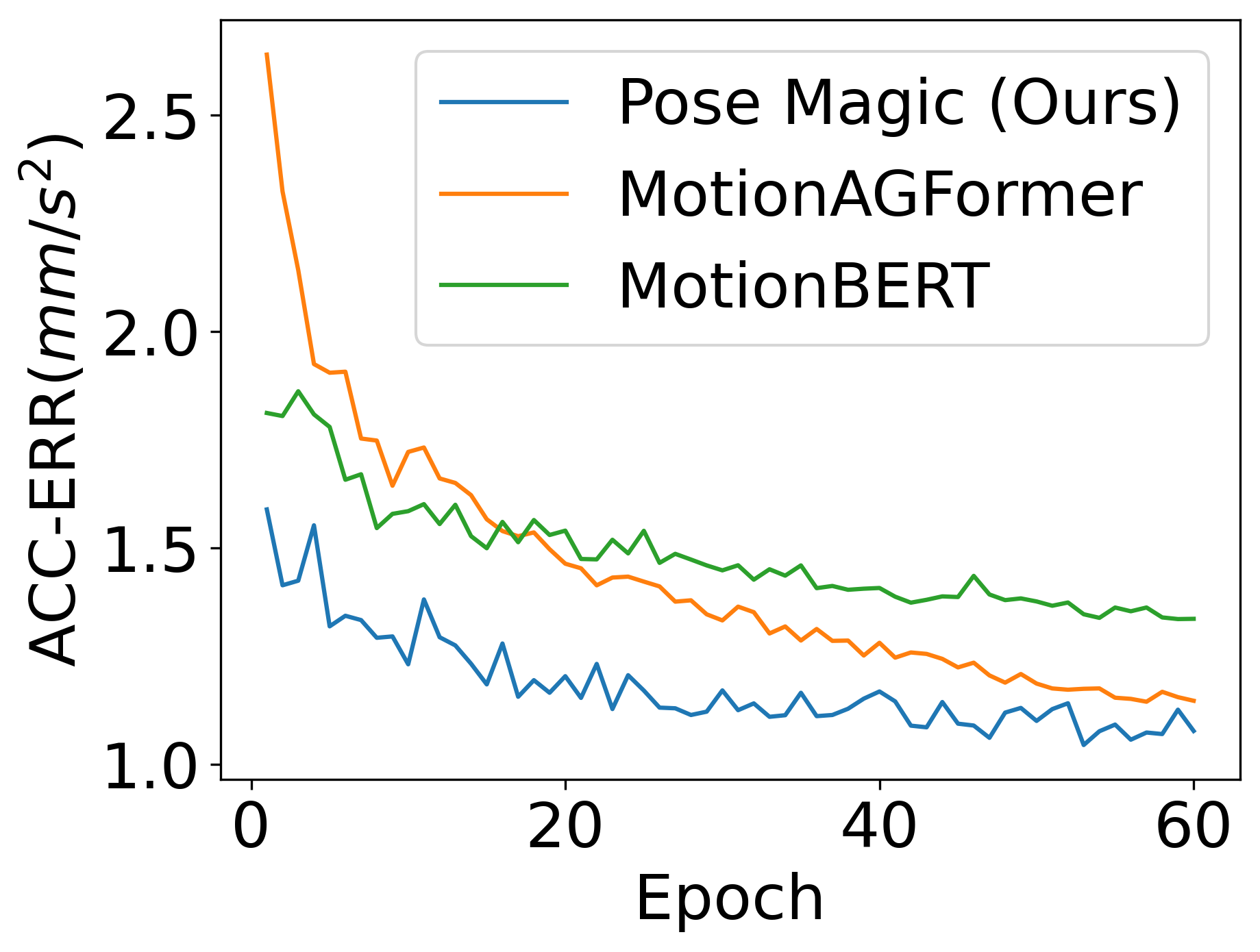}
        \centerline{(a) Causal models}
    \end{minipage}%
    \begin{minipage}[t]{0.49\linewidth}
        \centering
        \includegraphics[width=1.0\textwidth]{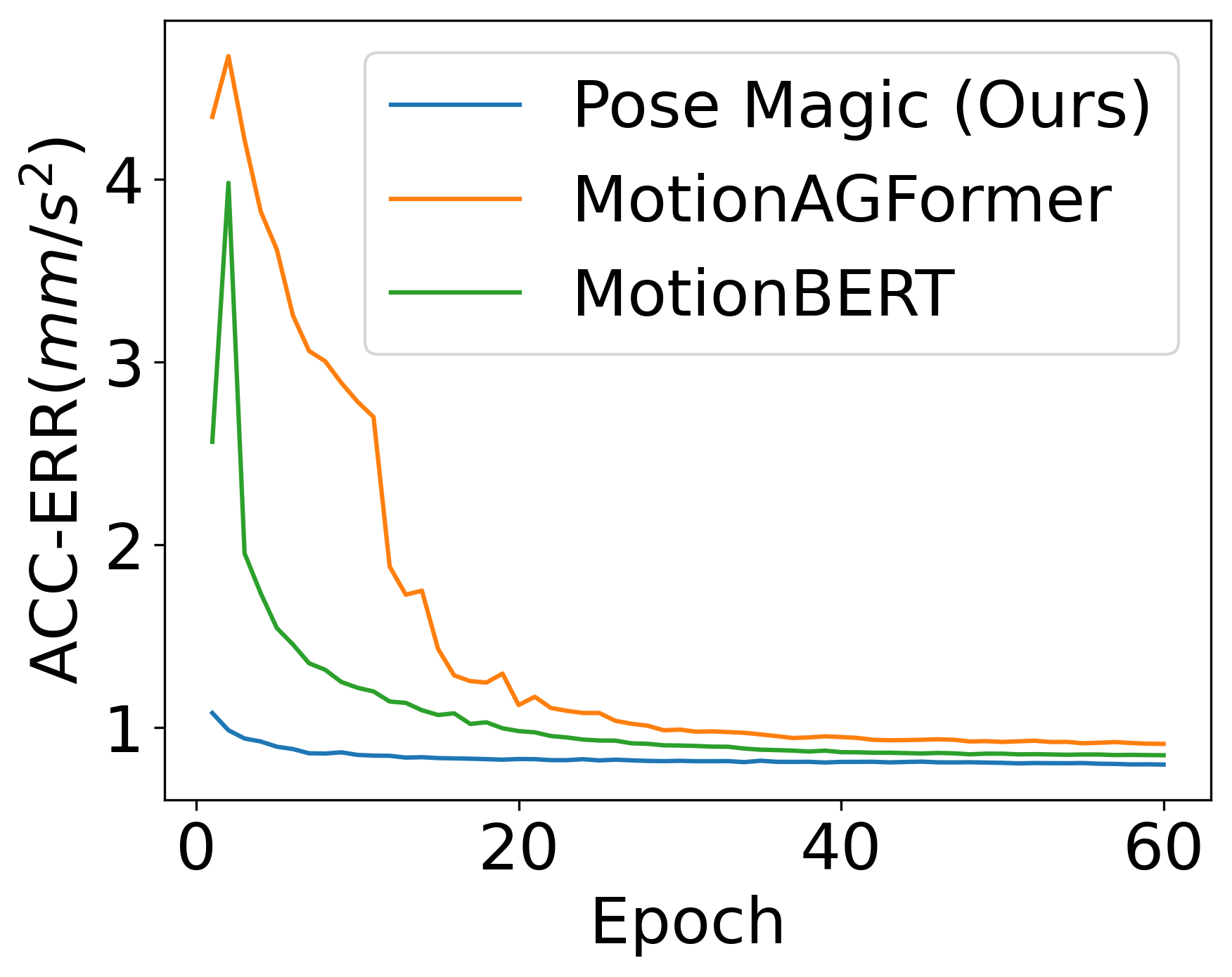}
        \centerline{(b) Bidirectional models}
    \end{minipage}
    \caption{Comparison results of ACC-ERR on Human3.6M.}
\label{ACC_ERR}
\end{figure}

We evaluate the ability to recover smooth 3D human motion from videos using MPJVE and ACC-ERR, as shown in Table \ref{mpjve} and Fig.\ref{ACC_ERR}. Our method achieves lower MPJVE and ACC-ERR, and converges faster. These results indicate that our method effectively models the natural smoothness of human motion by learning kinematic characteristics of human movement and current frame features in long-term relationships. We attribute this temporal coherence advantage to the inherent continuous-time of Mamba.

\subsection{Ablation Study}

\begin{table}[t]
\centering
\begin{tabular}{lcc}
\hline
Method & MPJPE$\downarrow$ &MPJVE$\downarrow$ \\
\hline
GCN only &54.2 &4.6\\
Mamba only &41.2  &1.8\\
GCN~$\rightarrow~$Mamba~(Sequential) &44.2 &1.8\\
Mamba~$\rightarrow$~GCN~(Sequential) &46.8  &3.9\\
Mamba~$\rightarrow$~GCN~(Parallel) &37.5 &1.6\\
\hline 
\end{tabular}%
\caption{Comparison of different integration. All models are trained on Human3.6M with bidirectional settings.}
\label{arche}
\end{table}

\textbf{Effectiveness of the Proposed Magic Block.} To verify the effectiveness of the proposed Magic Block, Table \ref{arche} shows alternative blocks. When only using GCN, MPJPE and MPJVE are $54.2mm$ and $4.6mm/s$, indicating its limited ability to accurately and smoothly capture the underlying 3D sequence structure. This is because GCN can only capture local dependencies. The combination of GCN and Mamba results in significant improvements. While local information is also available to Mamba, the parallel stream including GCN allows to balance the integration of local and global information. Compared to using Mamba alone, MPJPE and MPJVE are reduced by $3.7mm$ and $0.2mm/s$.

\begin{table}[t]
\centering
\begin{tabular}{lcc}
\hline
Method & MPJPE$\downarrow$ &MPJVE$\downarrow$ \\
\hline
No Embedding &37.8 &1.6\\
Spatial Embedding only &37.5 &1.6\\
Temporal Embedding only &38.0  &1.7\\
Both Embeddings &38.8 &1.7\\
\hline 
\end{tabular}%
\caption{Different types of positional embedding. All models are trained on Human3.6M with bidirectional settings.}
 \label{positional_embedding}
\end{table}

\textbf{Impact of Positional Embedding.} Table \ref{positional_embedding} explores the impact of positional embedding on accuracy and smoothness. It can be seen that incorporating temporal positional embedding on top of spatial positional embedding increases MPJPE by $1.3mm$. This is due to the non-permutation equivariant nature of GCN and the temporal continuity of Mamba. Unlike Transformer-based methods, our method inherently maintains the temporal sequence, thus eliminating the need for temporal positional embedding.

\begin{figure}[t]
    \begin{minipage}[t]{0.5\linewidth}
        \centering
        \includegraphics[width=1.0\textwidth]{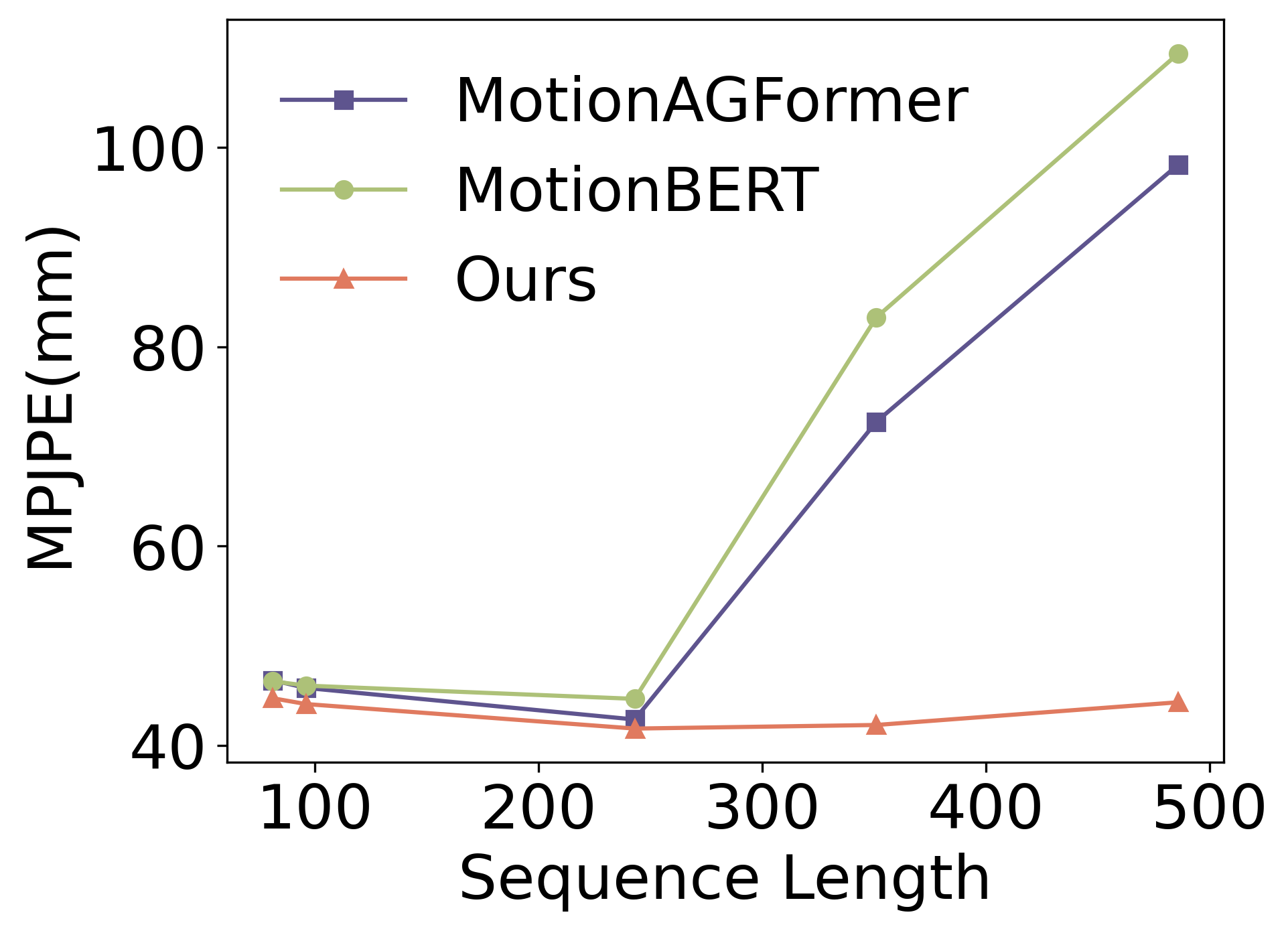}
        \centerline{(a) Causal models}
    \end{minipage}%
    \begin{minipage}[t]{0.5\linewidth}
        \centering
        \includegraphics[width=1.0\textwidth]{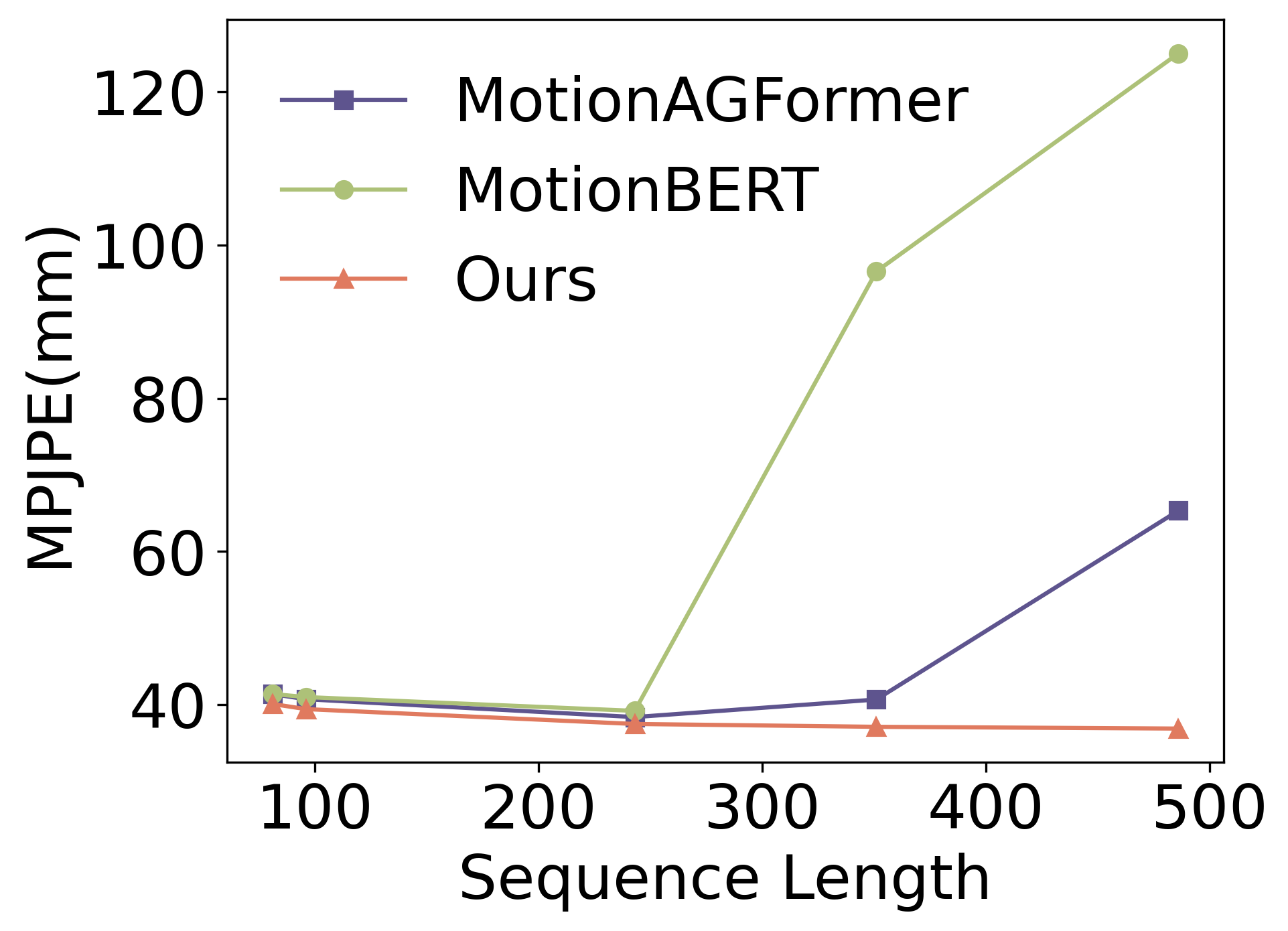}
        \centerline{(b) Bidirectional models}
    \end{minipage}
    \caption{Generalization performance on Human3.6M for different sequence lengths.}
\label{sequence}
\end{figure}

\textbf{Generalization to Unseen Sequence Length.} We further investigate the ability of Pose Magic to generalize to unseen sequence lengths. Pose Magic is initially trained on $T=243$ and then tested on $T=81, 96, 351,486$. As shown in Fig.\ref{sequence}, Pose Magic demonstrates consistent performance when test on shorter and longer sequences. Compared to MotionBERT \cite{zhu2023motionbert} and MotionAGFormer \cite{mehraban2024motionagformer}, Pose Magic can successfully generalize to any unseen sequence with minimal generalization loss ($\textless 3mm$), especially in encoding longer contexts. This is particularly beneficial for deployment scenarios where the sequence lengths do not match those used during training.

\section{Conclusion}
We propose a novel attention-free hybrid spatiotemporal architecture, Pose Magic, to address the trade-off between accuracy and efficient computation in 3D HPE. Specifically, we introduce the advanced state space model, Mamba, to effectively capture global dependencies. To complement this, we incorporate GCN to capture local joint relationships, enhancing neighborhood similarity and addressing local dependencies. This fusion improves the ability to understand the inherent 3D structure within input 2D sequences. Additionally, we provide a fully causal version of Pose Magic to perform real-time inference. Empirical evaluations demonstrate that Pose Magic achieves SOTA results while maintaining efficiency. Moreover, Pose Magic exhibits optimal motion consistency and smoothness, and can generalize to unseen sequence lengths.

\clearpage

\section{Acknowledgments}
This work was partly supported by the Special Foundations for the Development of Strategic Emerging Industries of Shenzhen (No.KJZD20231023094700001) and Zhejiang Provincial Natural Science Foundation of China (No.LQN25F010018).
\bibliography{aaai25}

\end{document}